\documentclass[conference]{IEEEtran}
\IEEEoverridecommandlockouts
\usepackage{cite}
\usepackage{amsmath,amssymb,amsfonts}
\usepackage{algorithmic}
\usepackage{graphicx}
\usepackage{textcomp}
\usepackage{xcolor}
\usepackage{subfig}
\usepackage{microtype}
\usepackage{url}
\usepackage{flushend}
\usepackage{tikz}
\usepackage{hyperref}

\newcommand\copyrighttext{%
  \footnotesize \textcopyright 2012 IEEE. Personal use of this material is permitted.
  Permission from IEEE must be obtained for all other uses, in any current or future 
  media, including reprinting/republishing this material for advertising or promotional 
  purposes, creating new collective works, for resale or redistribution to servers or 
  lists, or reuse of any copyrighted component of this work in other works. 
  DOI: \href{https://dl.acm.org/doi/10.5555/3523760.3523899}{10.5555/3523760.3523899}}
\newcommand\copyrightnotice{%
\begin{tikzpicture}[remember picture,overlay]
\node[anchor=south,yshift=10pt] at (current page.south) {\fbox{\parbox{\dimexpr\textwidth-\fboxsep-\fboxrule\relax}{\copyrighttext}}};
\end{tikzpicture}%
}

\def\BibTeX{{\rm B\kern-.05em{\sc i\kern-.025em b}\kern-.08em
    T\kern-.1667em\lower.7ex\hbox{E}\kern-.125emX}}
\begin{document}
 \setlength{\belowcaptionskip}{-10pt}

\title{EXOSMOOTH: Test of Innovative EXOskeleton Control for SMOOTH Assistance, With and Without Ankle Actuation*
\thanks{*Supported by the H2020 EUROBENCH project, grant No. 779963}
}

\author{\IEEEauthorblockN{Vittorio Lippi}
\IEEEauthorblockA{\textit{NZ} -
\textit{Uniklinik Freiburg}\\
Freiburg, BW Germany \\
0000-0001-5520-8974\\ vittorio.lippi@uniklinik-freiburg.de}\\
\IEEEauthorblockN{Francesco Porcini}
\IEEEauthorblockA{\textit{Institute of Mechanical Intelligence} \\
\textit{Scuola Superiore Sant'Anna}\\
Pisa, Italy  \\
0000-0001-9263-9423 }
\and
\IEEEauthorblockN{Alessandro Filippeschi}
\IEEEauthorblockA{\textit{IMI - DoE in Robotics and AI} \\
\textit{Scuola Superiore Sant'Anna}\\
Pisa, Italy \\
0000-0001-6078-6429}\\
\IEEEauthorblockN{Christoph Maurer}
\IEEEauthorblockA{\textit{Neurologie}  \\
 \textit{Uniklinik Freiburg}\\
 Freiburg, BW Germany \\
 0000-0001-9050-279X}
\and
\IEEEauthorblockN{Cristian Camardella}
\IEEEauthorblockA{\textit{Institute of Mechanical Intelligence} \\
\textit{Scuola Superiore Sant'Anna}\\
Pisa, Italy \\
0000-0002-3856-5731}\\
\IEEEauthorblockN{Lucia Lencioni}
\IEEEauthorblockA{\textit{Wearable Robotics S.r.L.} \\
Pisa, Italy }
}

\maketitle
\copyrightnotice

\begin{abstract}
This work presents a description of the EXOSMOOTH project, oriented to the benchmarking of lower limb exoskeletons performance.
In the field of assisted walking by powered lower limb exoskeletons, the EXOSMOOTH project proposes an experiment that targets two scientific questions. The first question is related to the effectiveness of a novel control strategy for smooth assistance. Current assist strategies are based on controllers that switch the assistance level based on the gait segmentation provided by a finite state machine. The proposed strategy aims at managing phase transitions to provide a smoother assistance to the user, thus increasing the device transparency and comfort for the user. The second question is the role of the actuation at the ankle joint in assisted walking. Many novel exoskeletons devised for industrial applications do not feature an actuated ankle joint. In the EXOSMOOTH project, the ankle joint actuation will be one experimental factor to have a direct assessment of the role of an actuated joint in assisted walking. Preliminary results of 15 healthy subjects walking at different speeds while wearing a lower limb exoskeleton supported the rationale behind this question: having an actuated ankle joint could potentially reduce the torques applied by the user by a maximum value of 85 Nm. The two aforementioned questions will be investigated in a protocol that includes walking on a treadmill and on flat ground, with or without slope, and with a load applied on the back.  In addition, the interaction forces measured at the exoskeleton harnesses will be used to assess the comfort of the user and the effectiveness of the control strategy to improve transparency.
\end{abstract}


\section{INTRODUCTION}
In the context of the lower limbs assistance during long lasting loads-carrying operations, a transparent and cost-effective exoskeleton has the potential to be the breakthrough for finally making those devices truly commercial.
In fact, if in the early years power assist exoskeletons were designed for a wide variety of applications, thus being bulky and expensive (e.g. \cite{mosher1968handyman,fontana2014body}), in the latest years there is a tendency towards a hardware simplification and a task-specific design of exoskeletons, e.g. \cite{kazerooni2019evaluation,zoss2006biomechanical,sesenna2021walking,wang2021active}, Ekso Vest (Ekso Bionics, Richmond, CA, USA), LegX  and suitX (US Bionics, Inc., Emeryville, CA, USA). In parallel, several control paradigms have been studied to improve the usability of these machines and cope with the limited sensing and actuation capabilities imposed by the hardware simplification. Therefore, it is nowadays fundamental to investigate the effects of hardware complexity and control strategies on the affordability and usability of exoskeletons in industry and logistics.
The EXOSMOOTH project, presented in this paper, aims at contributing to this investigation with a key application such as lower limb exoskeletons for load carrying. In particular, the investigation of hardware simplification will focus on the availability of a actuated ankle flexion module instead of a non-actuated one.

\begin{figure}[t!]
	\centering
		\includegraphics[width=0.90\columnwidth]{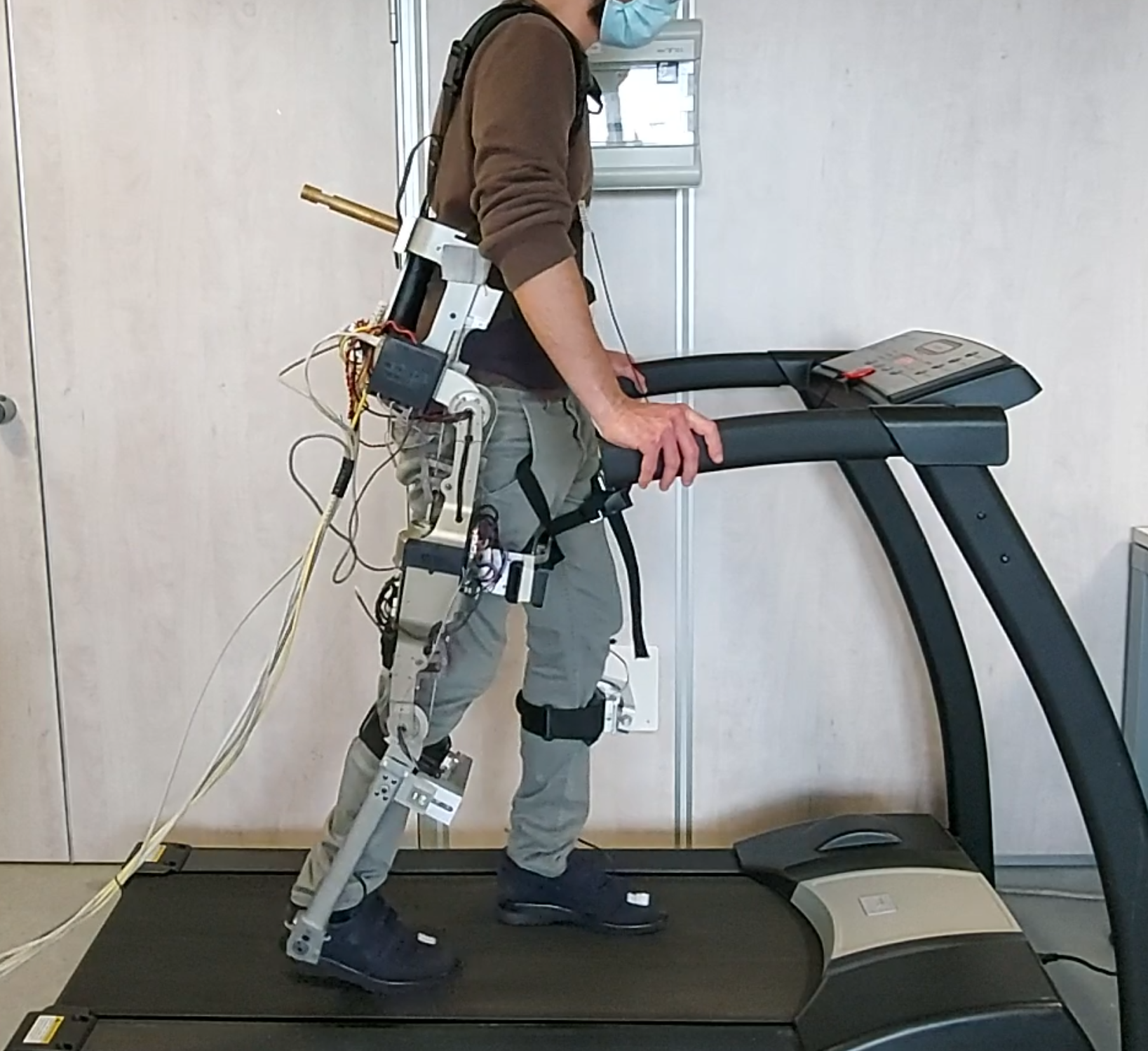}
	\caption{The Wearable Walker lower-limb exoskeleton. }
	\label{fig:Exoskeleton}
\end{figure}

\paragraph{Control strategies}
Current lower-limb exoskeletons control strategies involve EMG-driven, pre-defined gait trajectory, model-based, hybrid, and state-machine-based control loops \cite{yan2015review}. For those systems, the trade-off between the robustness and the quality of the control commands in the control loop is an important feature. Even when accurate trajectory tracking is achieved, the output torque is affected by discontinuities, caused by control signals shifting, based on gait phases or tracked features. This problem highly decreases the performance and usability of an assistive device. In the state-of-art, there are some attempts of solving the torque discontinuities problem applying a time-constant to the switching operation but this has not been tested across a consistent subjects’ population yet \cite{yan2015review,chen2019development}.

\paragraph{Ankle actuation}
Actuated ankle provides possibilities for advanced control and also to study balance control \cite{lippi2020challenge}. On the other hand, the possibility to reduce the number of actuated joints in a lower limb exoskeleton will lead to a cost reduction of the device itself, making it more advantageous on the market. Also, the improvement of the quality of the given assistance will make the overall system more reliable and safe from the usability point of view. Those factors will be a long step towards a higher diffusion of exoskeletons for supporting operators in industrial, lifesaving, and hostile environments, also shortening the time-to-market of future products.
Another important aspect of studying actuated ankle is its role in the overall system performance. Currently, no rigorous investigations have been conducted on the role of the actuated ankle joint in the transparency and overall usefulness in the control strategy. Most of the available exoskeletons do not include ankle actuation. Conversely human and humanoid posture control and balance studies highlighted the importance of ankle torque for standing \cite{peterka2002sensorimotor,T.Mergner2009,lippi2017human,lippi2016human, elverhoy2021ankle}. Comparing the two configurations (with and without ankle actuation) is a chance to clarify the role of the ankle joint and provide inspiration for the development of exoskeletons that are self-balancing or that aid balance. In particular, the transition between standing and walking and the transitions between different speeds (see the next section, “Experimental Protocol”) are expected to provide information about the “grey zone” between posture control (dominated by ankle strategy) and dynamic walking, where, in line of principle, ankle torque is less relevant.

\paragraph{The EXOSMOOTH project}
EXOSMOOTH is a subproject of the EUROBENCH European project (H2020 Framework - Grant 779963) which is oriented to the benchmarking of humanoids and exoskeletons in a perspective of real-world use \cite{torricelli2018eurobench, torricelli2020benchmarking,Lippi2019}.
The EXOSMOOTH project has been devised with the main motivation of testing an innovative control strategy to overcome issues related to discontinuous assistive torque and to investigate the role of ankle actuation in a relevant application such as load carrying. 
The main objectives of this project are: 1) Improvement of the state-of-art lower limb control strategy, through an innovative continuous-torque control \cite{Camardella2021} that eliminates the discontinuities of a state-machine based control, targeting higher transparency of the device and, thus, higher usability. 2) A study of the usefulness of ankle actuation in power augmentation exoskeletons from both transparency and assistance points of view, to reduce, if possible, the overall cost of a lower limb exoskeleton. 
Metrics of the project outcomes are the smoothness of assistive torque profiles, and transparency, assessed by measuring the interaction forces between the user and the exoskeleton.

The objectives will be attained through an experimental protocol that is presented in this paper along with a preliminary experiment on the current exoskeleton. 
In view of testing the effectiveness of the control strategy and the role of the ankle joint actuation in industrial settings, the experiment will include tasks such as carrying a load and moving on an ascending slope. Participants will wear the \textit{Wearable Walker} exoskeleton \cite{Camardella2021,icinco2021}, which is already available and will be updated with an actuated ankle joint, developed within the project itself. 

The experiment targets a preliminary estimation of the ankle torques needed to walk with the Wearable Walker exoskeleton. This provides a base to estimate the potential improvement of implementing ankle actuation, thus supporting the rationale of the experiment. 

\section{System Description}

\subsection{The Wearable-Walker Exoskeleton}
The lower-limb exoskeleton Wearable Walker (Fig. \ref{fig:Exoskeleton}) is an assistive device able to augment human capabilities to complete tasks such as load carrying. The exoskeleton consists of 7 rigid links connected by 8 joints: for each leg, there are two active joints (hip and knee flexion/extension) and two passive joints (ankle flexion/extension and ankle inversion/eversion). For the purpose of this work, an ankle module (with active flexion/extension and passive inversion/eversion) has been developed to be replaceable with the original passive ankle joint. This allows comparing the behavior of the exoskeleton with both active and passive ankle flexion/extension. The exoskeleton is altogether 110 cm high, 65 cm wide, and weighs 15 kg (off batteries). The exoskeleton's active joints are implemented by an actuation unit consisting of a brushless torque motor (RoboDrive Servo ILM 70x18) and a tendon-driven transmission system composed of a screw and
a pulley \cite{bergamasco2011high}. Each joint is provided with a Hall-effect sensor, while each motor shaft with an encoder. Moreover, the shoe insoles are sensorized with 4 pressure sensors and each leg featured 2 one-axis load cells, measuring human-robot force interactions at the thigh and the tibia. The power electronics group and the control unit are housed on the back-link of the exoskeleton. The power electronics group consists of a battery (75 V. 14 Ah) and a voltage conversion board, while the control unit consists of a computer running Simulink Real-Time at 5 kHz and a Wi-Fi communication module. A host PC is connected to the Simulink-Real Time PC through the Wi-Fi module.

\subsection{Blend Control Strategy}
The system controller is organized in three layers, that include a low-level layer (current control on each of the BLDC actuators and friction compensation), a middle-level layer (gravity and inertia balancing), and a high-level layer that orchestrates the middle-level layer components on each actuator according to the adopted control strategy, which includes the proposed Blend control.
The Blend control \cite{Camardella2021} is a novel control approach that provides smooth assistance torques. The algorithm is based on two separate Denavit-Hartenberg models, one refers to the left-foot-grounded (LFG) configuration and the other refers to the right-foot-grounded (RFG) configuration. These two models provide assistive torques which are "blended" together in order to obtain continuous and smooth overall assistance. This prevents the control from suffering torque discontinuities due to the model switching while walking. In particular, the overall assistive torque $\tau$ is defined as follows:

\begin{equation}
\tau = \gamma_{L}\tau_{L} + \gamma_{R}\tau_{R}
\end{equation}

where $\tau_{*}$ are the assistive torques and $\gamma_{*}$ are the blend gains in case of the LFG configuration ($*=L$) and RFG configuration ($*=R$). Both the assistive torques include a feed-forward dynamics and gravity compensation as well as friction and motor ripple compensation. Details about the calculation of $\tau_{*}$ are reported in \cite{Camardella2021}. The blend gains are defined as follows:

\begin{equation}
\begin{cases}
\gamma_{L}(q) = \dfrac{1}{2}(Y^{T}q + 1) \in [0,1] \subset \mathbb{R}\\
\gamma_{R}(q) = 1 - \gamma_{L}(q)
\end{cases}
\end{equation}

where $q$ is the joint angles vector and $Y$ is the constant gain matrix result of a linear regression. This linear regression aims at linking the joint angles vector with the stance phase (LFG or RFG) identified through the pressure sensors. More in detail, the regression is computed given the stance phase class (+1 LFG, -1 for RFG) as desired output and a joint angles vector time-series as input.

The middle layer components, i.e. gravity and inertia balancing, can be easily adjusted to account for either a known carried load or a known slope angle. In the EXOSMOOTH project both the load and the slope angle are known, but their measurement is foreseen as a future development of the system.

\section{Preliminary Experiment}

A preliminary validation of the system has been carried out in a controlled setting to assess the capability of the Blend control to achieve transparency. This experiment allowed the measurement of interaction forces at the harnesses and an estimation of the torque eventually needed at the ankle to obtain a full balancing of the Wearable Walker.

\subsection{Participants}
Fifteen male subjects (age $30.07\pm{4.50}$ years, height $1.75\pm{0.05}$ m,  weight $69.40\pm{4.65}$ kg), were enrolled for this study. All the participants signed a written disclosure and voluntarily joined the experiments. All the experiments have been conducted under the World Medical Association Declaration of Helsinki guidelines.

\subsection{Experimental set-up and protocol} \label{sec:exp}

Each participant was asked to walk under three velocity conditions while wearing the Wearable-Walker exoskeleton. In every speed condition, the Blend control strategy was employed and the ankle was never actuated. Data of the experiments were collected at 100 Hz by the host PC. All the PCs exchange UDP packets, for both syncing and data saving. 
The experiment includes low speed (T1), high speed (T3.5), and self-selected speed (SS) conditions. In particular, T1 refers to walking on the treadmill at 1 km/h for 30 s, T3.5 at 3.5 km/h for 30 s, and SS refers to ground walking for 7 meters.
First, subjects were asked to wear the exoskeleton and familiarize with the device (max 3 minutes). Then, participants ran a calibration phase in which collected data were used to train a gait phase classifier. 
Once trained the model, the experimental sessions were performed with the exoskeleton and the Blend control active, for each velocity condition. All the conditions, were randomized for each subject. 

\begin{figure*}[ht!]
     \centering
     \subfloat[][Ankles torque at 1 km/h speed]{\includegraphics[width=.33\textwidth]{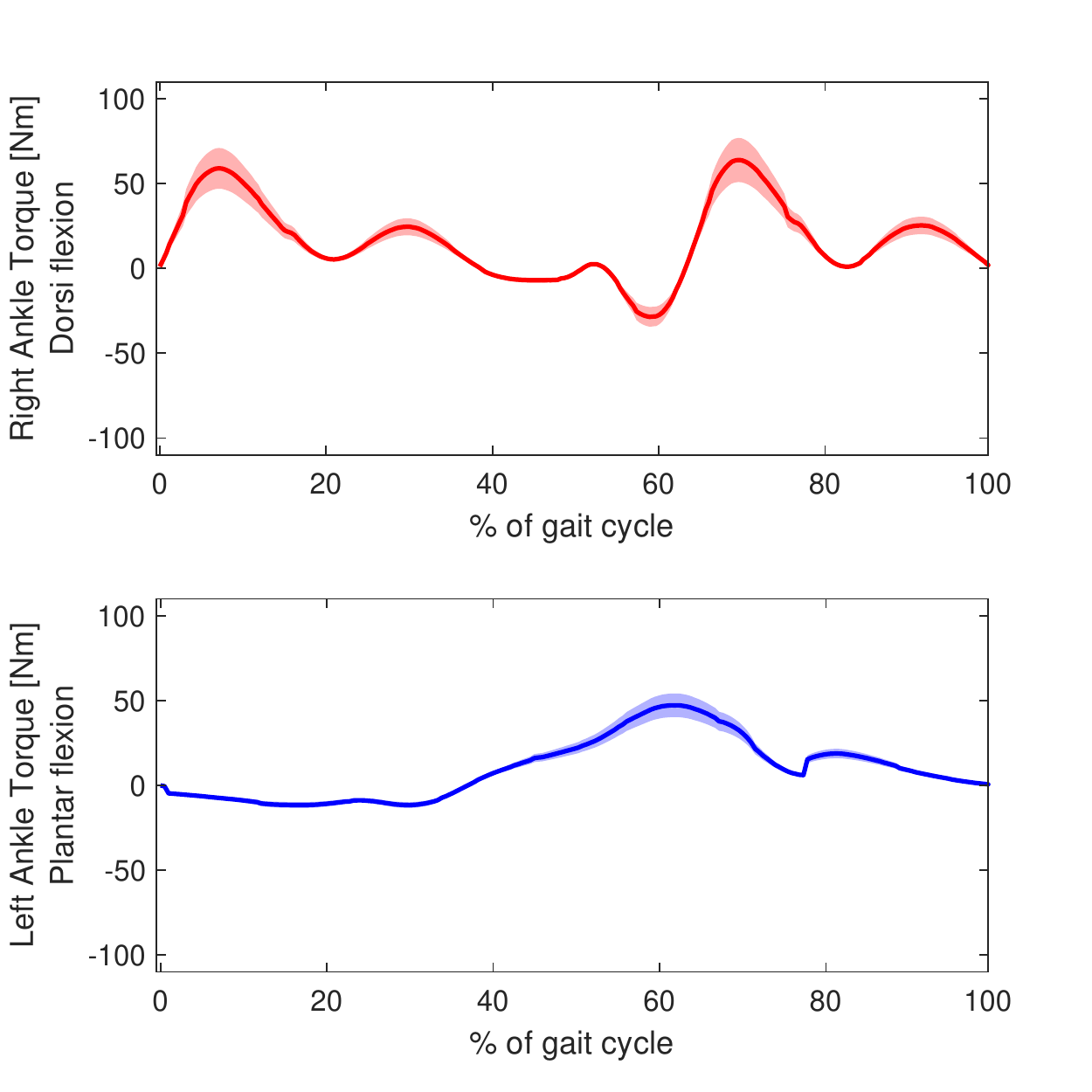}\label{fig:AT_1}} \hfill
     \subfloat[][Ankles torque at 3.5 km/h speed]{\includegraphics[width=.33\textwidth]{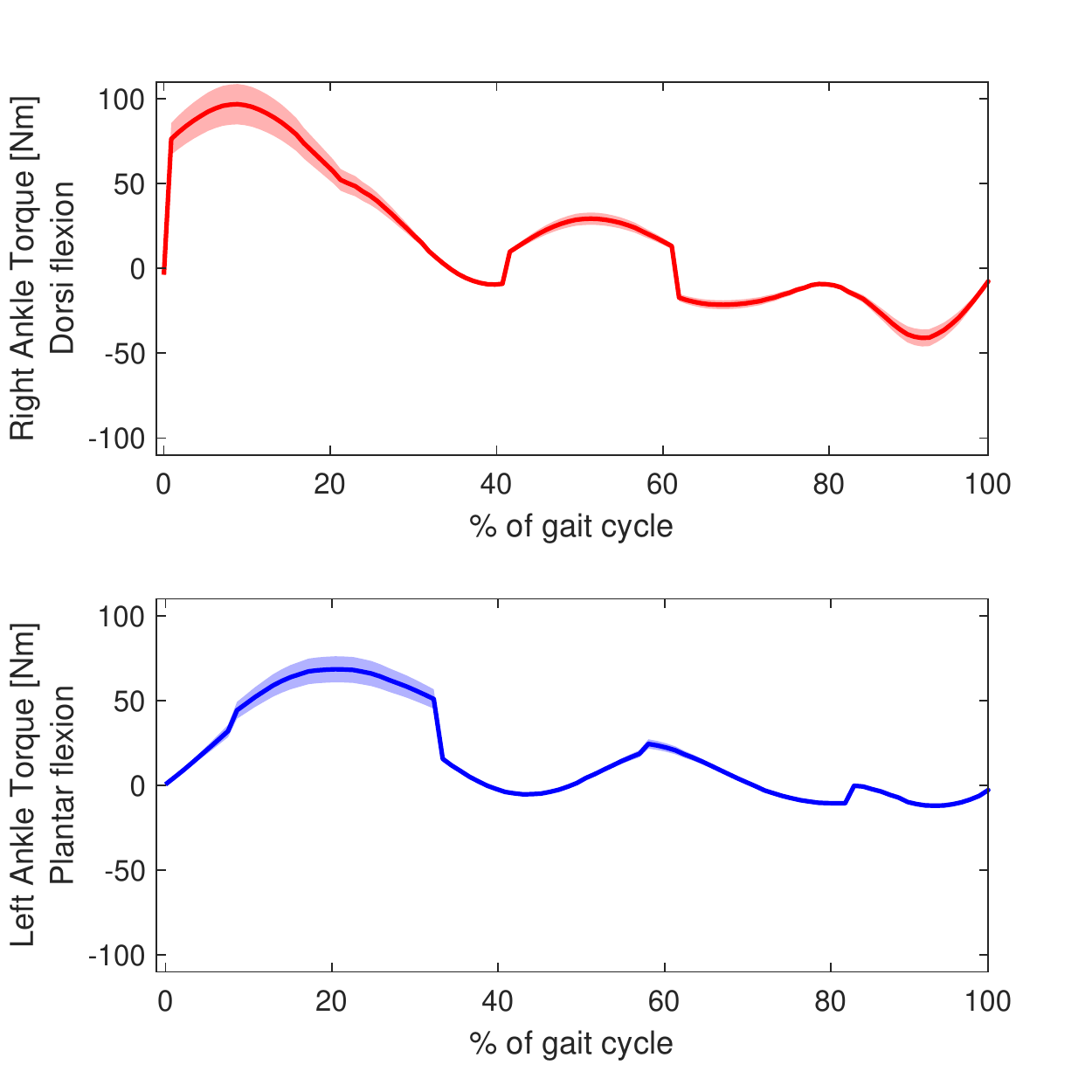}\label{fig:AT_35}} \hfill
		 \subfloat[][Ankles torque at self-selected speed]{\includegraphics[width=.33\textwidth]{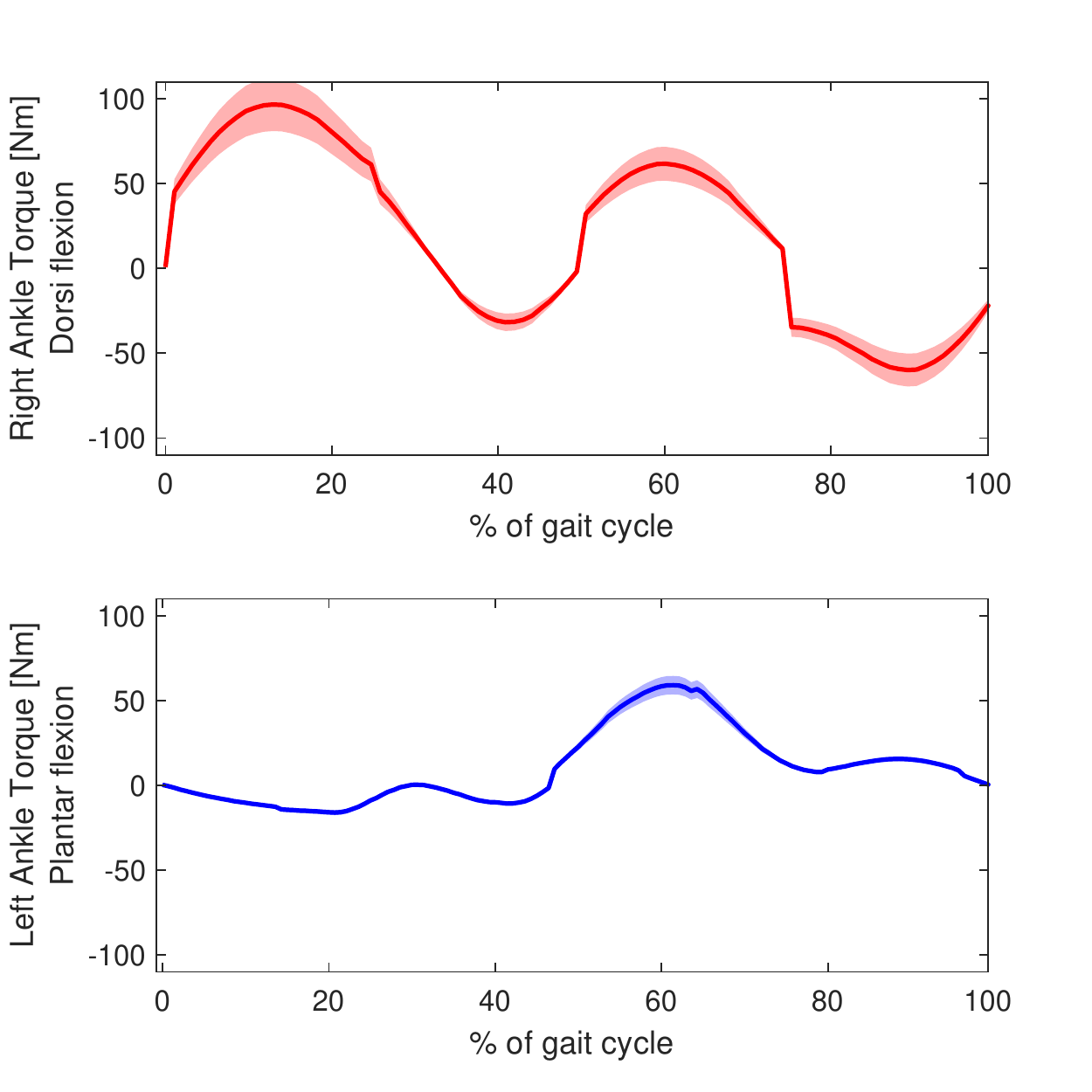}\label{fig:AT_ss}}
     \caption{Ankle torques with the three tested speed conditions. Colored bands show the inter-subjects torque peaks variation.}
     \label{fig:torquerespanel}
\end{figure*}

\subsection{Results}
The plots reported in Fig. \ref{fig:torquerespanel} show the torques that would have been commanded to the ankle actuators, if those joints were active, for gravity and inertial components compensation. Torque values refer to a generic gait cycle for both LFG (blue curve) and RFG (red curve) reference systems. The required torque increases as the walking speed increases, as it happens for physiological torques as well.
The torques required to the ankle joints follow a pattern in the gait cycle, and they are significantly bigger during the stance phase. In this phase, the peak balancing torque, which is mainly due to gravity compensation, reaches 72.15 Nm on average (18.43 Nm standard deviation) and the maximum value of 85 Nm.  These values are in the same order of magnitude of physiological torques reported in normal gait by people walking at a similar speed. In fact, Hansen et al. \cite{hansen2004human}, in an experiment that involved slow (0.90 m/s), normal (1.35 m/s), and fast (1.78 m/s) walking speed reported peak torques in the order of 1.7 Nm/kg. Novacheck in \cite{novacheck1998biomechanics} found smaller peaks in the order of 1 Nm/kg. Mueller et al. in \cite{mueller1995relationship} report a peak torque of 90.2 Nm (standard deviation 11.1 Nm) during walking at 1.09m/s (standard deviation 11.1 Nm). 
Moreover, in this experiment, the forces measured at the harnesses on the thighs and the shins are on average in the order of 25 N, with peaks of 80 N.

\begin{table}[]
    \centering    \begin{tabular}{c c c c}
    Speed & Average torque [Nm] & STD & Maximum torque [Nm] \\
    \hline
      1 km/h & 10.25  & 9.47  &  85.26  \\ 
    3.5 km/h  & 15.14  & 11.02  &  129.3 \\
    self selected   & 9.07 & 10.10  & 117.5\\
    \hline
    \end{tabular}
    \caption{Statistics of ankle torques in the three conditions.}
    \label{tab:protocol}
\end{table}

\section{Discussion and proposed protocol}
\subsection{Discussion}
The comparison of the literature with the experimental results shows that the user is required to supply an additional torque in the order of magnitude of the physiological one to balance the exoskeleton. This torque means an additional load to the user that is not negligible and that may affect balance, fatigue, and comfort. 
The additional torque needed to balance the exoskeleton is not necessarily provided in the proximity of the ankle joint:  all harnesses can be exploited by the user to balance the exoskeleton, and the redundancy of the forces that the user can exert on the exoskeleton makes it difficult, if not impossible, to predict their distribution. For these reasons it is important to devise and carry out experimental activities to understand the role of an active ankle joint, and to evaluate whether the eventual benefits are worth the bigger costs.
The magnitude of the forces exerted on the harnesses is a measure of the effects of ankle actuation. The obtained values show a potential for improvements: an ideally transparent device would allow the user to walk without exerting forces on the harnesses. Therefore, an overall reduction of the forces and a redistribution among the harnesses is expected as a consequence of the ankle actuation. At the same time, the measurement of the EMG activity on the leg muscles (e.g. gastrosoleus, anterior tibiae, hamstrings, and rectus) is expected to provide insights on the redistribution of the muscular activity, with a hypothesis of an overall reduction of EMG features related to the articular torques (e.g. root mean square of the signal).

\subsection{EUROBENCH experimental protocol}
The proposed experiments target the aforementioned objectives by including three factors: the hardware configuration, the assistance strategy, and the carried load. The first two factors are directly related to the goals of the study. The latter aims at investigating the effects of assistance amplification on the other factors. For example, will the advantages of continuous assistive torque be more evident when the carried load increase?
The hardware configuration includes three levels, i.e. no exoskeleton, exoskeleton without actuation at the ankles, and exoskeleton with actuation at the ankles. The assistance strategy includes two levels, i.e. assistance based on a finite state machine (FSM) for gait segmentation and the blended assistance strategy. The carried load has two levels as well, i.e. no load and a 10 kg load brought on the back, rigidly attached to the exoskeleton.
In addition, three conditions will be investigated: walking on a treadmill, walking on the ground, and walking on the treadmill with an ascending slope. These conditions are adapted from the EUROBENCH scenarios PEPATO and EXPERIENCE \cite{eurobenchws}.
The treadmill conditions allow for full control over the walking speed and the slope, whereas walking on flat ground is the condition that is closest to a real scenario.
The whole protocol is reported in table \ref{tab:protocol2}, in which each row is an experimental trial. The duration of each trial and number of repetitions will be tuned in pilot experiments before carrying out the experiments in the EUROBENCH facilities in Madrid.

\begin{table}[]
    \centering    \begin{tabular}{c c c c c}
    Ground & Slope & Load & Speed sequence [km/h] \\
    \hline
    TM   & no  & no  &  0$\rightarrow$2$\rightarrow$4$\rightarrow$6$\rightarrow$2$\rightarrow$6 \\ 
    TM   & yes  & no  &  0$\rightarrow$4 \\
    TM   & no  & yes  &  0$\rightarrow$2$\rightarrow$4$\rightarrow$6$\rightarrow$2$\rightarrow$6 \\
    TM   & yes  & no  &  0$\rightarrow$4 \\
    TM   & no  & no  &  4\\
    TM   & no  & yes &  4\\
    FG   & no  & no  &  fixed gait rhythm\\
    FG   & no  & yes &  fixed gait rhythm \\
    \hline
    \end{tabular}
    \caption{Experimental conditions for the eight trials. TM stands for treadmill and FG stands for flat ground. The arrow ``$\rightarrow$'' represents a gradual transition between the two speeds that takes 15 seconds. In the FG trials the gait rhythm will be suggested by means of a metronome.}
    \label{tab:protocol2}
\end{table}

\section{CONCLUSIONS}
This paper has presented a project aimed at the study of a novel control strategy and the role of ankle actuation for a lower limb exoskeleton. The project will include extensive experimental activities in tasks that are representative of real scenarios. In the paper, the room for improvements has been demonstrated through experimental activities and discussed in order to extract the research questions behind the experiment.

\bibliographystyle{ieeetr}

\bibliography{exosmooth_pres}

\end{document}